\documentclass{article}
\usepackage{times}
\usepackage{soul}
\usepackage{url}
\usepackage[hidelinks]{hyperref}
\usepackage[utf8]{inputenc}
\usepackage[small]{caption}
\usepackage{amsmath}
\usepackage{amsfonts}
\usepackage{algorithm}
\usepackage{graphicx}
\usepackage{float}
\usepackage{bm}
\usepackage{amsmath}
\usepackage{tikz}
\usepackage{ijcai22}
\usepackage[noend]{algpseudocode}
\DeclareMathOperator*{\argmax}{argmax}

\newcommand{\cross}[1]{
    \begin{scope}
    \coordinate (A) at (#1);
    \draw[thick] (A)+(-0.1,0) --(A)+(0,0) --(A)+(0,0.1) --(A)+(0,0) --(A)+(0.1,0) --(A)+(0,0) --(A)+(0,-0.1) --(A)+(0,0) --(A)+(-0.1,0) --(A)+(0,0);
\end{scope}
}
\newcommand{\agent}[1]{
    \begin{scope}
    \coordinate (A) at (#1);
    \filldraw[fill=cyan, draw=blue] (A) circle (0.4+rand*0.3);
    \filldraw[fill=yellow, draw=blue] (A) circle (0.2);
\end{scope}
}

\title{Improving generalization to new environments and removing catastrophic forgetting in Reinforcement Learning by using an eco-system of agents}
\author{
Olivier Moulin$^1$
\and
Vincent Francois-Lavet$^1$\and
Paul Elbers$^2$\And
Mark Hoogendoorn$^1$
\affiliations
$^1$Vrije Universiteit Amsterdam\\
$^2$Amsterdam UMC
}

\begin{document}
\maketitle
\begin{abstract}
Adapting a Reinforcement Learning (RL) agent to an unseen environment is a difficult task due to typical over-fitting on the training environment. RL agents are often capable of solving environments very close to the trained environment, but when environments become substantially different, their performance quickly drops. When agents are retrained on new environments, a second issue arises: there is a risk of catastrophic forgetting, where the performance on previously seen environments is seriously hampered. This paper proposes a novel approach that exploits an eco-system of agents to address both concerns. Hereby, the (limited) adaptive power of individual agents is harvested to build a highly adaptive eco-system. 
\end{abstract}

\section{Introduction}
\label{Introduction}
In Reinforcement Learning, over-fitting to the environment on which the agent has been trained is a common problem (cf. \cite [Cobbe et al., 2019] {KCobbe2019}), resulting in poor performance in unseen environments that are substantially different from the one trained upon.
\begin{figure}[ht]
	\centering
    \begin{tikzpicture}
        \draw[thick] (-3.5,-1.7) --(-3.5,2.7) --(1.8,2.7) --(1.8,-1.7) --(-3.5,-1.7);
        \agent{0,0};
        \cross{0,0};
        
        \agent{0,2};
        \cross{0,2};

        \agent{-2.5,2};
        \cross{-2.5,2};

        \agent{-2.7,-1.1};
        \cross{-2.7,-1.1};

        \agent{1,-0.9};
        \cross{1,-0.9};

        \agent{-1,-1};
        \cross{-1,-1};

        \agent{-2,1};
        \cross{-2,1};

        \agent{-2,0};
        \cross{-2,0};

        \agent{0.75,0.75};
        \cross{0.75,0.75};

        \agent{-1,0.8};
        \cross{-1,0.8};

        \foreach \x in {0,...,80}{
          \cross{-1+rand*2,0.5-rand*2}}
        \cross{-3.5,-2}
        \node(crossnode) at (-2,-2){Environment};
        \filldraw[fill=yellow, draw=blue] (-0.3,-2) circle (0.2);
        \node(tasklearned) at (1.2,-2){Task learned};
        \filldraw[fill=cyan, draw=blue] (-2.3,-2.75) circle (0.2);
        \node(tasklearned) at (-0.4,-2.75){Adaptability};
    \end{tikzpicture}
	\caption{Our approach: eco-system adaptability concept}
	\label{adaptability_concept}
\end{figure}
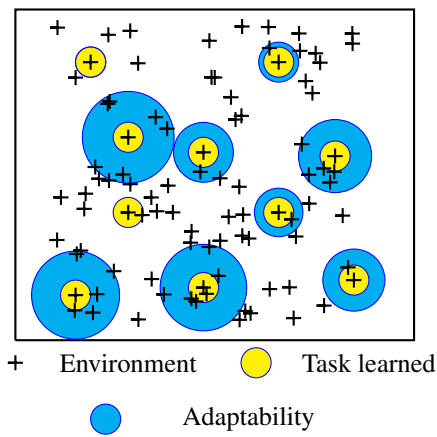
This is particularly challenging in cases where agents are regularly confronted with new environments.
Generalization to never seen or modified environments has been defined in Packer et al., 2018 \cite {CPacker2018}.
A variety of approaches have been introduced to address this challenge. The approaches proposed in the literature often focus on a single agent being trained on several environments and trying to learn a higher-level strategy to solve this set of environments (see e.g.\ \cite[Sonar et al., 2018]{asonar2020}, \cite[Igl et al., 2019]{Migl2019}). These approaches are typically shown to have better performance on multiple environments.
While such existing approaches certainly improve performance on generalizability, they have several important downsides. First of all, they typically suffer from so-called catastrophic forgetting, which means that the agent looses the capability to perform well in environments it has explored in details / solved in the past. This issue has been detailed and solutions have been proposed by continuing to sample experiences from past environments when training on new ones (cf.\ in \cite[Atkinson et al., 2020] {catkinson2020}), but this does not fully remove the catastrophic forgetting risk. The second downside is that retraining on new environments can be very expensive, especially in cases where limited computational resources are available.
In this paper, we propose an approach based on the observation that agents trained on an environment can generalize to environments that are close to the original training environment (Figure \ref{adaptability_concept}).
We combine a set of such agents in an eco-system, where each agent is trained on one specific environment. This means that we have a whole set of agents, each with a limited capacity to generalize, grouped as an eco-system to increase the group capacity to generalize to a rich set of environments. The eco-system expands by adding new agents when new environments are encountered that cannot be solved by the current eco-system, which means that each agent in the pool is never retrained after being added to the pool. We evaluate the approach in two different environments, one being the minigrid framework and the other based on a newly defined setting. We also benchmark our approach against an existing approach (cf. \cite [Igl et al., 2019] {Migl2019}) to assess generalization increase compared to a simple voting ensemble technique (cf. \cite [Wiering et al., 2008] {MAWiering2008}).
This paper is organized as follows. Section \ref{related_work} presents the related work. Section \ref{approach} follows by an explanation of our approach. Next, Section \ref{experimental_setup} presents the experimental setup used to evaluate the approach. Section \ref{results} presents the results of the experiments and compares our approach with the chosen benchmarks. We end with a discussion in Section \ref{discussion_conclusion}.
\section{Related work}
\label{related_work}
\subsection{Generalization}
We defined generalization capability or generalizability of a given system as its capacity to perform properly on never seen environments after being trained on a set of other environments.
A significant number of papers have looked at generalization capacity of a reinforcement learning agent to never seen environments. It has been defined by Packer et al. \cite {CPacker2018} and assessed on a set of environments and RL algorithms by Cobbe et al. \cite{KCobbe2019}. Mutlitple of approaches have been presented to improve generalization in RL,  out of which we can see two major tracks.
The first category of approaches focus on trying to create a representation of the environment to learn a good policy with a more limited risk of over-fitting. An example of this work can be found in the paper from Sonar et al. \cite{asonar2020} where the researchers propose a new algorithm in which the agent learns a representation on which the action predictor is optimal on all the training environments. These approaches, although they improve the generalizability of the agent by reducing over-fitting to an environment, still suffer from catastrophic forgetting.
The second category of approaches focus on adding noise and/or information bottlenecks in the learner (e.g. neural network) to avoid over-fitting to the environment. This approach is for instance used by Chen \cite{jzchen2020} by implementing a Surprise Minimization algorithm, which is an algorithm that provides an estimated reward based on probability according to the history of rewards by state. 
A similar approach is used by Lu et al. \cite{Xlu2020} and Igl et al. \cite{Migl2019} where the researchers propose to add an information bottleneck between the environment and the agent as well as including noise to avoid over-fitting. These types of approaches, have proven to be successful on reducing over-fitting, but again do not solve the catastrophic forgetting problem.
\subsection{Catastrophic forgetting}
To address catastrophic forgetting, most of the literature focuses on three alternatives: (1) making sure the agent retrains on previous tasks (e.g. \cite[Atkinson et al.,  2020]{catkinson2020}); (2) modifying weights used for a past task only slowly (cf. \cite[Kickpatrick et al., 2016]{JKickpatrick2016}), or (3) keeping several versions of the neural network with a progressive network architecture (e.g. \cite[Rusu et al.,  2016]{AARusu2016}). These approaches significantly reduce catastrophic forgetting but cannot completely avoid it, which is the case for our proposed approach. Our approach solve this problem at the expense of an additional computational resource cost. This cost tends to reduce and sometime even get smaller than the other approaches when the eco-system grows.
\section{Approach}
\label{approach}
Our eco-system approach is based on the assumption that each agent trained on an environment embeds some capacity to generalize, which allows it to perform well on a limited number of other environments. 
The eco-system is composed of  multiple agents (a pool of agents).
Each agent being part of the eco-system is trained as a standard Reinforcement Learning agent.
\subsection{Reinforcement Learning formulation}
We consider an agent interacting with its environment over discrete time steps.
The environment is formalized as an MDP defined by (i) a state space $\mathcal S$ that is possibly continuous, (ii) an action space $\mathcal A=\{1, \ldots, N_{\mathcal A}\}$, (iii) the transition function $T:~\mathcal S \times \mathcal A \times \mathcal S \to [0,1]$, (iv) the reward function $R:~\mathcal S \times \mathcal A \times \mathcal S \to \mathcal R$ where $\mathcal R$ is a continuous set of possible rewards in a range $R_{\text{max}} \in \mathbb{R}^{+}$ ($[0,R_{\text{max}}]$).
An MDP $M$ starts in a distribution of initial states $b_0(s)$.
At time step $t$, the agent chooses an action based on the state of the system $s_t \in \mathcal S$ according to a policy $\pi:\mathcal S \times \mathcal A \rightarrow [0,1]$.
After taking action $a_t \sim \pi(s_t,\cdot)$, the agent then observes a new state $s_{t+1} \in \mathcal S$ as well as a reward signal $r_t \in \mathcal R$.
\subsubsection{The meta-learning setting}
In this paper, we do not consider the task of learning in a single environment but a distribution of different (yet related) environments $M \sim \mathcal M$ that differ in the reachable states from the state space $\mathcal S$ while the action space $\mathcal A$, the reward function $R$ and the transition function $T$ are the same (see Section \ref{experimental_setup} for two examples).
This type of setup matches the need of a system to adapt to small changing tasks, like for example a new labyrinth (like in the minigrid environment used for this paper) or a new environment for a robot in charge of doing a specific task.
The agent aims at finding a policy $\pi(s,a)$ with the objective of maximizing its expected return, defined (in the discounted setting) as
$$V^\pi(b_0)=\underset{M_i \sim \mathcal M}{\mathbb{E}} \left[ \sum \nolimits_{k=0}^{\infty} \gamma^{k} r_{t+k} \mid s_t=s, \pi  \right].$$
\\where 
$$r_{t} = \underset{a \sim \pi(s_t,\cdot)}{\mathbb E} R \big(s_{t},a, s_{t+1} \big)$$
and 
$$\mathbb P \big( s_{t+1} | s_{t}, a_t \big)=T(s_{t},a_t,s_{t+1})$$.
In our approach, we consider that an agent has "solved an environment" when the total reward collected $\mathcal R$ is bigger than a predefined threshold $l$, even if it is not the optimal total reward. This threshold is defined to be what is considered a good enough policy to behave properly in an environment. This threshold changes based on the environment presented to the system. 
\\The Double Deep Q Network (DDQN) algorithm and the Proximal Policy Optimization algorithm (PPO) have been used for running the experiments (detailed in \ref{experimental_setup}). Using these two different algorithms showed that the improvements brought by the eco-system approach is independent from the optimization algorithm used.
\subsubsection{Q-learning}
In Reinforcement Learning, the agent learns a policy (sequence of actions) which allows it to get the best reward in its environment.
In order to leverage all past experiences, we use an off-policy learning algorithm that samples transition tuples $(s, a, r, \gamma, s')$ from a replay buffer.
The Q-function is learned using the Double Deep Q-Network (DDQN) algorithm (cf. \cite [Van Hasselt et al., 2019] {HVanHasselt2019}), which uses the target:
    $$Y = r + \gamma Q (s', \argmax_{a' \in \mathcal{A}} Q(s', a'; \theta_{Q}); \theta_{Q^-})$$
where $\theta_{Q^-}$ are parameters of an earlier buffered Q-function (or our target Q-function). The agent then minimizes the following loss:
    $$L_{Q}(\theta_{Q}) = (Q(s, a; \theta_{Q}) - Y)^2$$
From this Q-value function we can derive the policy $\pi$ learned by the agent as
	$$\pi(s)=  \underset{a \in \mathcal A}{\argmax}\, Q(s, a; \theta_{Q}), \forall s \in \mathcal{S}$$
The specific version of DDQN used in this paper is based on DeeR implementation \cite [Francois-Lavet, 2016]{VFrancoisLavet2016}.
\subsubsection{PPO}
The Proximal Policy Optimization (PPO) method (cf. \cite [Schulman et al., 2017] {jschulman2017}) is an extension of the actor-critic method where the parameters $w$ of the policy $\pi_{w} (s,a)$ are updated to optimize $A^{\pi_{w}}(s,a)=Q^{\pi_{w}}(s,a)-V^{\pi_{w}}(s)$. The PPO algorithm also enforces a limit on the policy changes, which results in maximizing the following objective in expectation over $s~\sim~\rho^{\pi_w}, a~\sim~\pi_w$:
$$\min \Big( r_t(w) A^{\pi_w}(s,a), \text{clip} \big(  r_t(w), 1-\epsilon, 1 + \epsilon \big) A^{\pi_w}(s,a)  \Big)
$$
\begin{itemize}
    \item $r_t(w)=\frac{\pi_{w+\bigtriangleup w} (s,a)}{ \pi_{w} (s,a)}$,
    \item $\rho^{\pi_{w}}$ is the discounted state distribution, which is defined as
\\$\rho^{\pi_{w}}(s)=\sum_{t=0}^{\infty} \gamma^t Pr\{s_t=s | s_0, {\pi_{w}} \}$, and
    \item $\epsilon \in \mathbb R$ is a hyper-parameter.
\end{itemize}
The specific version of PPO used in this paper is based on stable baselines 3 (cf. \cite [OpenAI]{sb3PPO}).
\subsection{Agent eco-system}
Our approach, based on an eco-system of agents, makes use of the generalizability of each agent to cover a higher number of environments in a given distribution of environments $\mathcal{M}$ (Figure \ref{adaptability_concept}).
An agent is considered trained on a given environment when it can reach a reward threshold $l \in \mathbb R$. The policy learned is then considered as the satisfactory policy to solve the environment.
\\Our algorithm works as follows: new environments are drawn sequentially and randomly from the distribution $M_i \sim \mathcal M$, $i~\in~\mathbb N$. When the first environment is drawn, there is no agent in the pool. 
When a new environment $M_{i}$ is presented to the algorithm, it first looks whether one agent $e_{j}$ from the pool $\mathcal{E}=\{e_1, \ldots, e_n\}$ can solve $M_{i}$ without additional training. 
When a new environment is presented to the eco-system, the policy $\pi$ of a given agent $e_{j}$, noted as $\pi^{e_{j}}$, will be tested on the new environment. In the worst case, each agent $e_{j}$ being part of the pool of agents $\mathcal{E}$ need to be checked. 
Our approach assumes that it is not an issue to initially try out many policies, but that training a new one/retraining an existing one is a bigger problem and should be reduced to minimal amount. (for example in a limited computational setup like IOT)
Our approach tries to replace as much as possible the learning phases by inference phases which consumes far less computational resources.
In order to optimize the search of a good enough policy $\pi^{e*}$ in the pool of agents $\mathcal{E}$, the agents in the pool are ordered by number of environments they have already solved, thus checking the policy $\pi^{e_{j}}$ that solved the highest number of environments first.
If $e_{j}$ can solve $M_{i}$ then the new environment is added to the list of environments the agent $e_{j}$ can solve. Otherwise, a new agent $e_{n+1}$ is created, trained on $M_{i}$ until it can reach the reward threshold $l$ and is added to the pool $\mathcal{E}$. 
This new agent $e_{n+1}$ is then tested on the list of environments previously solved by the pool  $\mathcal{E}$, and each environment solved by $e_{n+1}$ is added to its list of environments.
If $e_{n+1}$ can solve all the environment of an existing $e_{j}$ agent in the pool $\mathcal{E}$ then $e_{j}$ is removed from the pool.
This part is optional as generalizability will work without it, but this ensure that the pool of agents $\mathcal{E}$ is kept to a reasonable number of agents $e$ at the expense of having to test new agents on many environments. This setting is needed in a system with limited resources (memory, computational capabilities) in order to keep the pool of agents manageable.
This approach can be discussed, as it creates an additional inference workload each time a new agent is added to the pool. However, by replacing less effective agents by a smaller number of better agents, we also reduce the test time of the agents in the pool each time an unknown environment is presented, which we have observed happens more often than adding a new agent to the pool. Also this optimization part is as indicated above not mandatory to ensure good performance of our approach and it is only composed of inference and not training actions, which are quite low on the computational side.
The pesudo-code of our approach is detailed in Algorithm~\ref{codetrainingmultiagent} with $e_{n}$ being the agent $n$ of the pool $\mathcal{E}$, $\mathcal{R}^{\pi_{e_{n}}}_{M_{i}}$ being the total reward gathered by the agent $e_{n}$ on the environment $M_{i}$, $l$ representing the threshold to consider the policy $\pi^{e_{n}}$ successful, $e^{*}$ identifying the agent which solved successfully the new environment, and $\delta^{e_{n}}$ being the list of environments solved successfully by the agent $e_{n}$. 
\begin{algorithm}[ht]
	\caption{eco-system - learn($M_{i})$} 
	\label{codetrainingmultiagent}
    \begin{algorithmic}
    \State $e^{*}$ $\leftarrow$ $\emptyset$ \textit{\scriptsize \ \#good enough agent found}
	\State $n$ $\leftarrow$ 0 \textit{\scriptsize\ \#loop var.}
	\While{$e^{*}$ = $\emptyset$ and $\bigcup$ $e_{0...n}$ $\neq$ $\mathcal{E}$} 
        \State \textit{\scriptsize\ \#while good policy not found }
		\State \textit{\scriptsize\ \#and not all agents reviewed}
		\State $\mathcal{R}^{\pi_{e_{n}}}_{M_{i}}=$ \text{test\_agent}($e_{n}$,$M_{i}$)
        \State \textit{\scriptsize \ \#Total reward from $e_{n}$ on $M_{i}$}
		\If {$\mathcal{R}^{\pi_{e_{n}}}_{M_{i}}$ $\geq$ $l$ } \textit{\scriptsize \#if $e_{n}$ solve $M_{i}$}
			\State $e^{*}$ $\leftarrow$ $e_{n}$ \textit{\scriptsize \ \#good enough agent found = $e_{n}$}
		\Else 
            \State $n$ $\leftarrow$ $n+1$
        \EndIf
	\EndWhile
	\If {$e^{*}$ = $\emptyset$} \textit{\scriptsize \#if $e^{*}$ not found}
		\State	$e$ $\leftarrow$ new\_agent()
		\While{$\mathcal{R}^{\pi_e}_{M_{i}}$ $\leq$ $l$} \textit{\scriptsize \ \#while $e$ cannot solve $M_{i}$}
		    \State learn-epoch($e$,$M_{i}$) 
            \State $\mathcal{R}^{\pi_e}_{M_{i}}=$ \text{test\_agent}($e$,$M_{i}$)
	    \EndWhile
		\State $\mathcal{E}$ $\leftarrow$ $\mathcal{E}$ + $e$ \textit{\scriptsize \ \#add $e$ to the pool}
		\State \textit{\scriptsize \ \#Following For statement is optional}
		\State \textit{\scriptsize \ \#Only needed if optimization of the pool is needed}
		\For {$f$ $\in$ $\mathcal{E}$} \textit{\scriptsize \#for all agent $f$ in the pool}
			\For {$w$ $\in$ $\delta^{f}$} \textit{\scriptsize \#for all env. $w$ solved by $f$}
				\State $\mathcal{R}^{\pi_{e^{*}}}_{w}=$ \text{test\_agent}($e^{*}$,$w$)
                \If {$\mathcal{R}^{\pi_{e^{*}}}_{w}$ $\geq$ $l$} \textit{\scriptsize \#if $e$ can solve $w$}
				    \State $\delta^{e}$ $\leftarrow$ $\delta^{e}$+$w$ \textit{\scriptsize \#add $w$ to $e$ list}
				\EndIf
			\EndFor
			\If {$\delta^{f}$ $\in$ $\delta^{e}$ }
				\State \textit {\scriptsize \#if $e$ can solve all env. of $f$}
				\State $\mathcal{E}$ $\leftarrow$ $\mathcal{E}$ - $f$ \textit{\scriptsize \#remove $f$ from pool}
			\EndIf
		\EndFor
		\State Sort $\mathcal{E}$ by size $\delta$ descending order
	\Else
		\State $\delta^{e^{*}}$ $\leftarrow$ $\delta^{e^{*}}$+$M_{i}$ \textit{\scriptsize \ \#add $M_{i}$ to $e^{*}$ list}
	\EndIf		
    \end{algorithmic}
\end{algorithm}
Each agent of the pool is only trained on one environment and never retrained on anything else during its life in the pool, ensuring that catastrophic forgetting is not an issue.
The distribution $\mathcal{D}$ of tasks successfully solved by the eco-system, given the distribution of environments $d(e)$ successfully solved by the agent $e$ in the pool $\mathcal{E}$ (seen or not already seen), is defined as:
$$\mathcal{D}=\bigcup_{e=0}^{n}d(e), e \in \mathcal{E}$$
Essentially, the eco-system behaves as a knowledge accumulator, meaning that it continues expanding its knowledge and optimizes its storage, while ensuring it maintains agents to solve at least all the environments it has been previously trained on.
\section{Experimental setup}
\label{experimental_setup}
\subsection{Environments}
\begin{figure}[ht]
	\centering
	\includegraphics[width=0.5\linewidth]{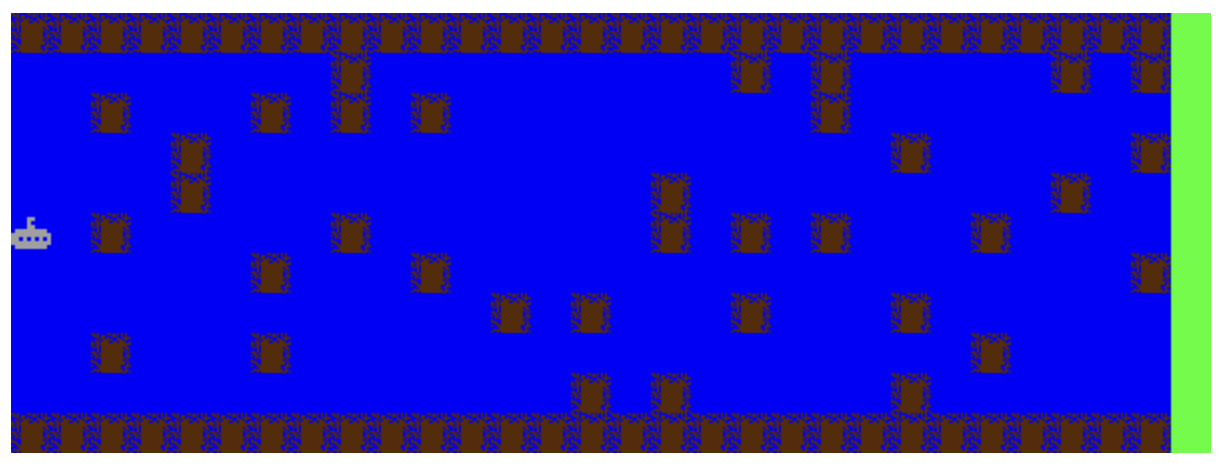}
	\caption{Submarine environment}
	\label{submarine}
\end{figure}

\begin{figure}[ht]
	\centering
	\includegraphics[width=2.5cm]{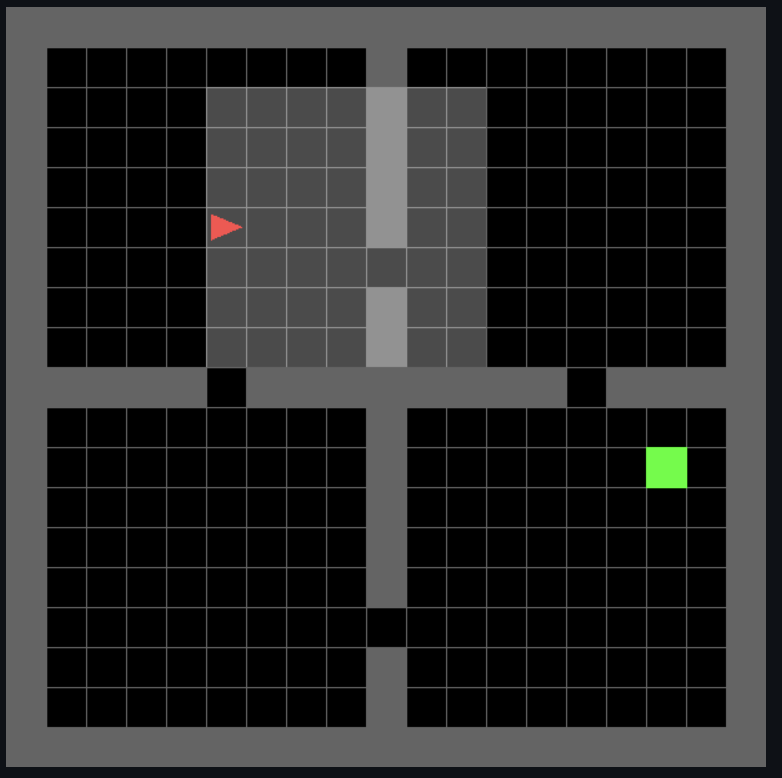}
		\includegraphics[width=2.5cm]{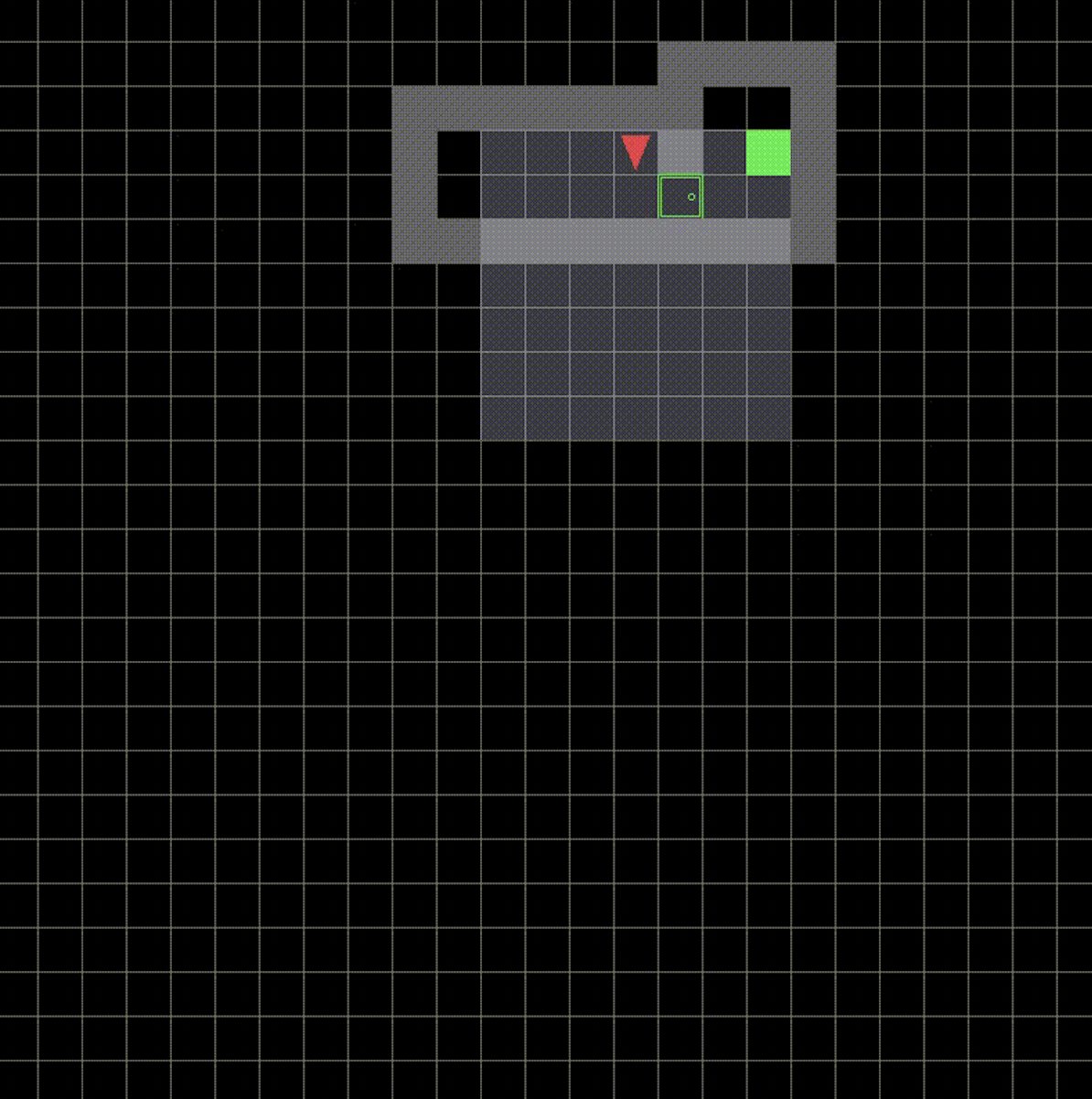}
	\caption{Minigrid FourRoom and Multiroom environment}
	\label{minigrid}
\end{figure}

The experiments have been conducted on two different environments (a newly developed Submarine environment and Minigrid), each having two different settings. 
\paragraph{Submarine} First, the Submarine environment (Figure \ref{submarine}) is introduced for this paper. 
In this environment, the agent controls a submarine which is moving from left to right by one column after each step. The agent can choose between 3 actions (up, stay, and down) and should reach the right side without hitting any block. The blocks are randomly placed at each level using a seed (Figure \ref{submarine}). Hitting a block or reaching the right side ends the game, with a -100 reward in the first case and +100 in the second one.
This is a simple environment (with only 3 actions) while still being a challenging environment (all obstacles are randomly placed at each level). The observations can be easily customized to check how our approach performs on different settings.
Two versions of the Submarine environment were created. The first one, \emph{Submarine-easy}, has a state representation that allows for a better generalizability which includes the current y position of the agent as well as a view of the next 5 columns $[x:x+5]$ and all rows in these columns $[0:11]$.
This state is a pattern which can be easily found in multiple environments, which helps reducing over-fitting and increase the probability that the output given by the neural network in an environment for this state will also work on another environment.
The second version, \emph{Submarine-hard}, makes it more difficult to learn a general strategy, which  leads to smaller generalizability, as the state is composed of the x and y position of the agent as well as the next 15 columns $[x:x+15]$ and all rows in these columns $[0:11]$. 
\paragraph{Minigrid} On top,  we use Minigrid with two environments, namely the FourRoom and the  MultiRoom environments (cf. \cite[Chevalier-Boisvert et al., 2018] {gym_minigrid}). 
We have chosen this Minigrid setting to match the approach made by Cobbe et al. \cite {KCobbe2019} using procedurally generated environments.
In our case, the map, start position, goal position, and obstacles are positioned randomly according to the seed of each level, while the other components of the experiments like reward given to the agent are kept the same for each level.
In order to benchmark our approach we have chosen to compare to the paper from Igl et al. \cite{Migl2019} as it performs very well on the Minigrid environment. We run experiments on the FourRoom environment  (Figure \ref{minigrid}) and the MultiRoom environment (Figure \ref{minigrid}). The MultiRoom environment was used in the experiment of the benchmarking paper.
In these environments the basic view of Minigrid is used as state and the agent gets a reward for reaching the goal in the maze. The reward is defined as :
$$1 - 0.9 * (stepsUsed / maxStepsAllowed)$$
\subsection{Baselines}
The baselines (single agent, IBAC-SNI introduced by Igl et al. \cite{Migl2019}) are chosen to match the approaches used in papers listed in the related work section: a single agent trying to generalize to new environments.
\paragraph{Submarine setup}For the Submarine environment the performance analysis was made between a single agent on one side and the eco-system on the other side trying to solve 1000 never seen environments after having been trained on 1000 other environments using the DeeR \cite{VFrancoisLavet2016} framework.
The Submarine environment experiments have also been run using stable-baselines3 and PPO to ensure that the results were not linked to a particular algorithm.
The choice has been made to use the exact same code, used for training each agent in the eco-system, to train the single agent approach in order for the performance increase of the eco-system can be easily identified.
We have decided not to compare to IBAC-SNI on this environment has we wanted to compare to the benchmarking paper only on environments used in it (Minigrid), to ensure a fair comparison.
The ensemble voting technique has been run on the Minigrid FourRoom environment to assess its performance.
\paragraph{Minigrid setup}For the Minigrid FourRoom and MultiRoom environments, the performance analysis was made between the code associated with the paper from Igl et al. \cite{Migl2019} and the eco-system after being trained for $10^{8}$ steps (as used in IBAC-SNI paper) on different environments randomly chosen, using the stable-baselines3 framework and the PPO learning algorithm.
On top, a comparison has been made for the Minigrid FourRoom environment, as it was the most challenging one from both Minigrid environments, between the eco-system and our implementation of the vorting ensemble technique described by Wiering et al. \cite{MAWiering2008} and indicated as the most efficient ensemble technique they have tested. 
\paragraph{Ensemble voting technique}
One baseline was created to compare the performance of the eco-system to the Reinforcement Learning voting ensemble technique described by Wiering et al. \cite{MAWiering2008}.
To initialize the ensemble technique, we have used 200 environments and trained one agent on each of them and added the agents to a pool.
During the training, every 20 environments, we have tested the pool of agents (composed of 20 agents, then 40, then 60 ...) on 1000 never seen environments and computed the average of total reward gathered.
For testing the pool of agents on each environment, the following approach was taken: at each step on each environment, each agent in the pool was voting for which action to take. The majority vote is then taken.
\paragraph{Pool ordering impact}
We have also tested, in the Minigrid FourRoom Environment, if changing the order for testing the agents in the pool has an effect on the results, by ordering the agent in the descending order of solved environments and then on the ascending order.
\subsection{Training algorithms}
In the experiments conducted, different learning algorithms (Double DQN, PPO) and frameworks (DeeR, stable-baselines3) have been used.
\paragraph{Threshold definition}
The threshold for reward is noted $l$, in the following formulas, but it will differ based on the type of environment used, for example $l_{s}$ is the threshold for the Submarine environment.
The threshold used is the same for all environments of a given type.
\paragraph{Submarine environment}
The threshold $l_{s}$ has been set to 100 for the Submarine game as any total reward $\geq$ 100 means that the agent has reached the goal. The threshold $l_{m}$ has been set to 0.8 for the Minigrid environments (1 for reaching the goal minus a penalty taken for each step). 
\paragraph{Minigrid environment}
In order to define the Minigrid threshold at 0.8, we have assessed if changing this threshold has a direct impact on the generalization result of the experiment. We have tested 3 different threshold (0.8, 0.85 and 0.9). The results are indicated in section \ref{results}.
More details about the technical implementation, and hyper-parameters are indicated in the appendix of this paper.
\subsection{Performance metrics}
In order to asses the performance of each approach, 3 main metrics are considered : the generalizability index, the catastrophic avoidance index and the amount of access to the environments.
\paragraph{The generalizability index} This metric is based on testing on a number of unseen environments $M_{i} \in \mathcal{M}$. It expressed whether an approach can perform over a certain threshold $l$ or what average of total reward $\mathcal{R}$ is gathered by the approach.
The \emph{generalizability index based on a threshold}, noted as $\omega$, is indicated as a percentage representing the percentage of times the approach has performed over the threshold $l$ on all the new environments $M_{i} \in \mathcal{M}$ on which it was tested. It can be noted as:
\begin{center}
	$\omega = \frac{\sum_{i=0}^{n} \begin{Bmatrix}
			R_{M_{i}}>=l \rightarrow  1
			\\ R_{M_{i}}<l  \rightarrow 0
		\end{Bmatrix}*100}{n}$
\end{center}
The emph{generalizability index based on the average reward gathered}, noted as $\zeta$, is indicated as a float value, showing the average of total rewards $\mathcal{R}$ gathered over all the new environments $M_{i} \in \mathcal{M}$ on which the approach was tested. It is formalized as follows:
\begin{center}
	$\zeta = \frac{\sum_{i=0}^{n} R_{M_{i}}}{n}$
\end{center}
 \paragraph{The catastrophic forgetting avoidance index} This metric, noted $\xi$, is based on testing periodically during the initial training how each approach performs on the environment $M_{i} \in \mathcal{M}$ on which it has already been trained. It checks that the approach does not forget the knowledge already accumulated. The metric is indicated as a percentage representing the percentage of time the agent $e$ total reward $\mathcal{R}$ was over the threshold $l$ on previously learned environments. It can be noted as:
\begin{center}
	$\xi = \frac{\sum_{i=0}^{n} \begin{Bmatrix}
			R_{M_{i}}>=l \rightarrow  1
			\\ R_{M_{i}}<l  \rightarrow 0
		\end{Bmatrix}*100}{n}$
\end{center}
\paragraph{The amount of access to the environments}: this metric is based on testing periodically how much access (number of states observed) has been made to the environment by each approach. This is a fair comparison as it accounts for the number of access made during learning on a given environment but also the number of access used during inference when testing if a given agent can solve an environment.
The metrics are calculated periodically after each approach has been presented to 50 additional environments (and completed the associated training if necessary).
\section{Results}
\label{results}
\subsection{Submarine environment}
\begin{figure}[ht]
	\centering
	\includegraphics[width=5cm]{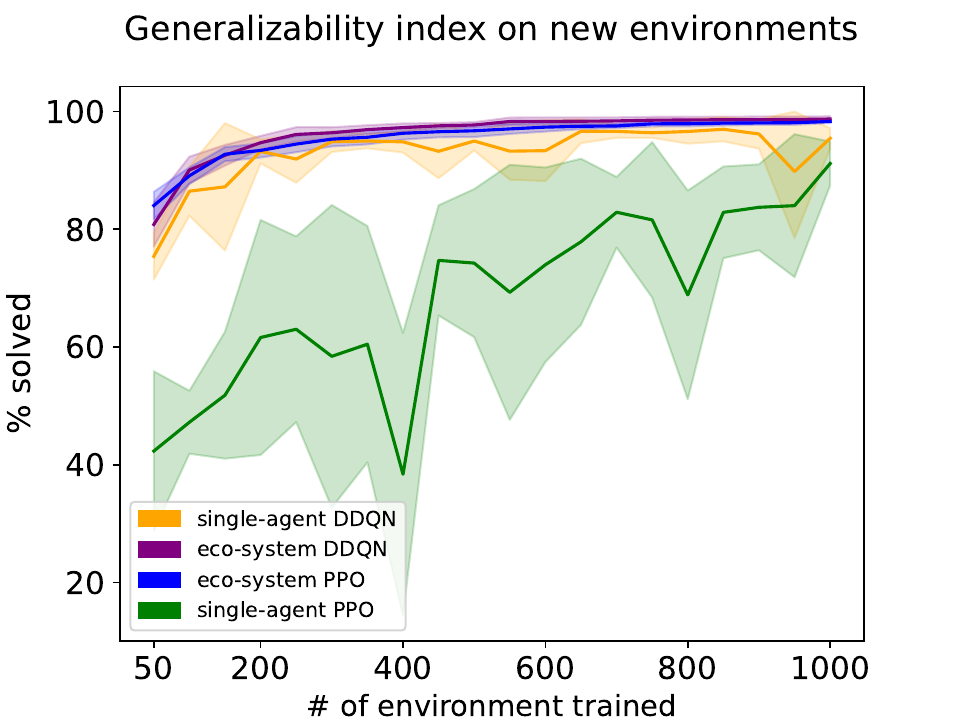}
	\caption{Generalization Submarine easy}	
	\label{easy_generalization_gen}
\end{figure}

\begin{figure}[ht]
	\centering
	\includegraphics[width=5cm]{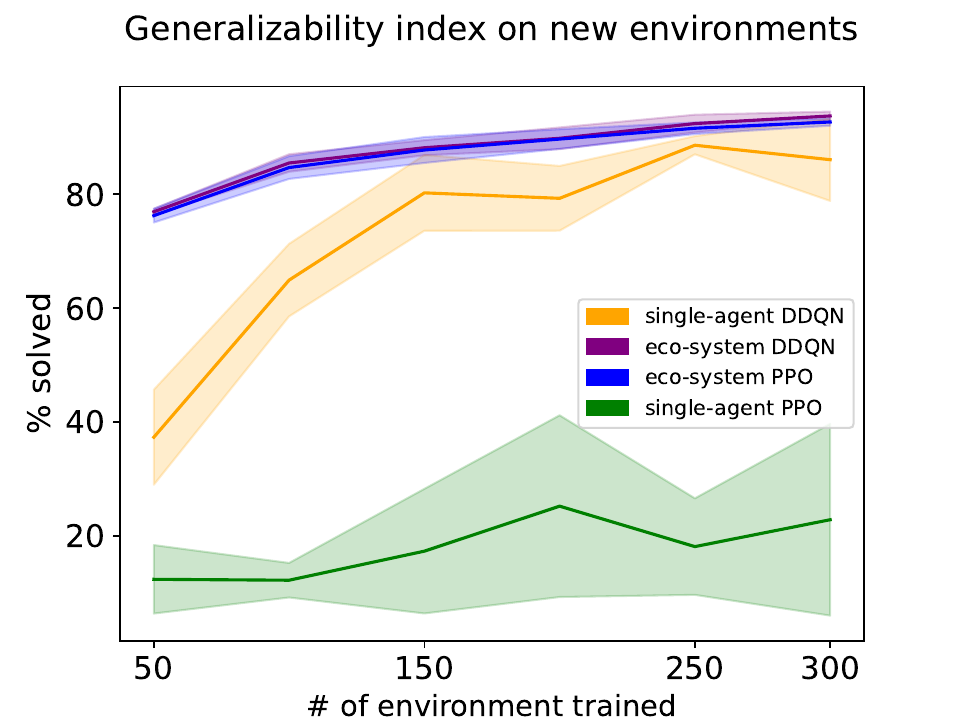}
	\caption{Generalization Submarine hard}
	\label{hard_generalization_gen}
\end{figure}

\begin{figure}[ht]
	\centering
	\includegraphics[width=5cm]{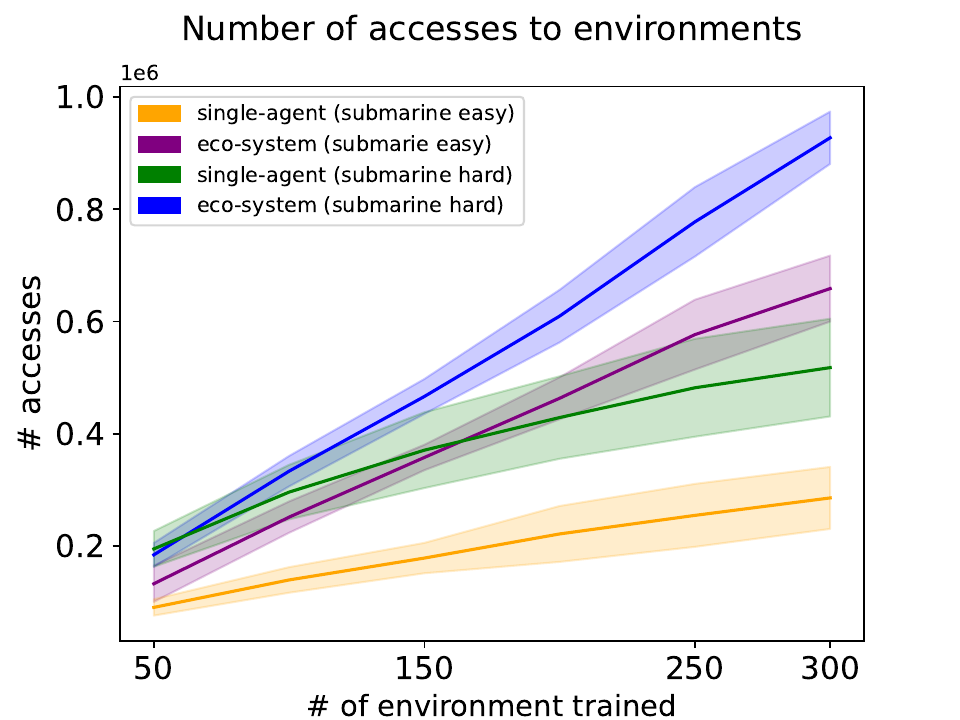}
	\caption{Number of accesses to the environment}
	\label{hard_access}
\end{figure}

\begin{figure}[ht]
	\centering
	\includegraphics[width=5cm]{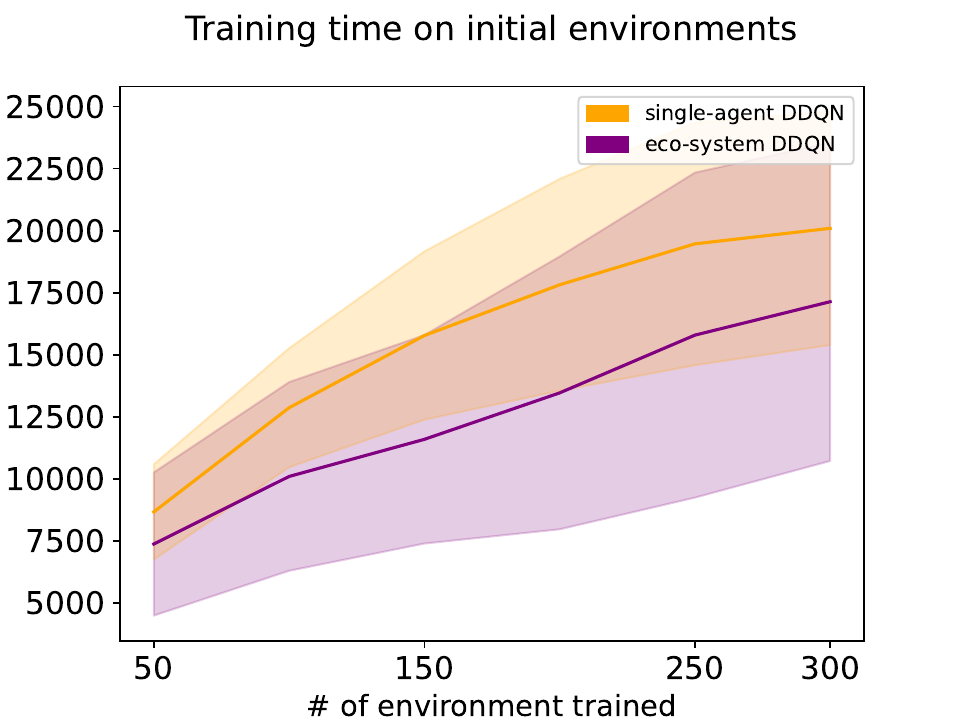}
	\caption{Training duration comparison}
	\label{hard_duration}
\end{figure}

\paragraph{Submarine-easy} The first set of results has been run on the "Submarine-easy" environment.
For the generalizability index based on threshold $\omega$, considering a distribution of 1000 new environments for testing (Figure \ref{easy_generalization_gen}), the eco-system is far more stable than the one agent approach and also gives a better result at the end of the training.
\begin{figure}[ht]
	\centering
	\includegraphics[width=5cm]{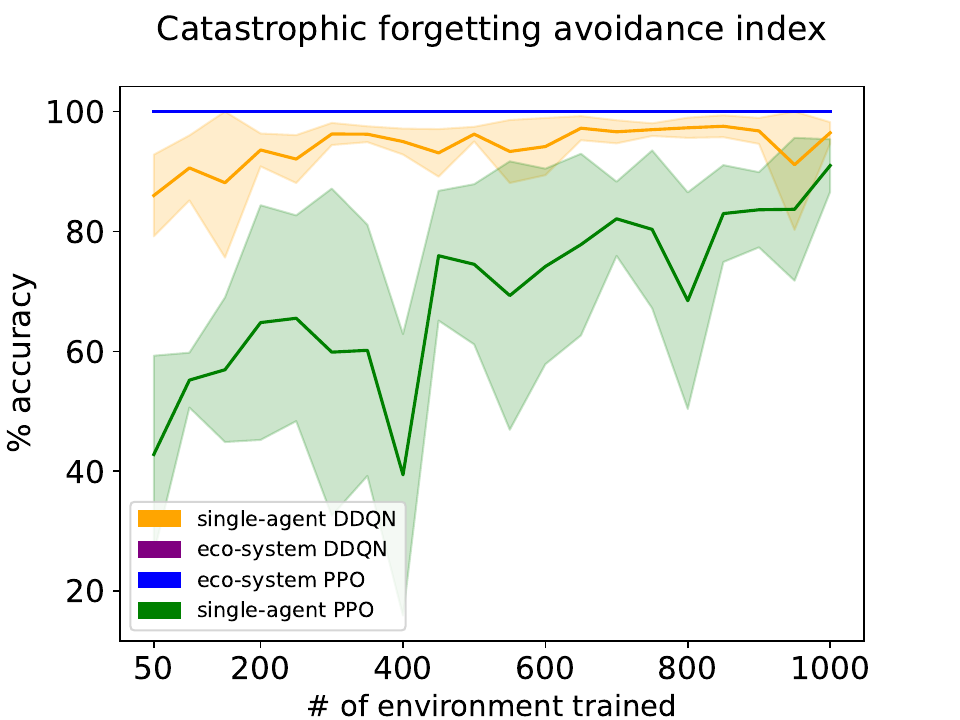}
	\caption{Catastrophic forgetting Submarine easy}
	\label{easy_generalization_forget}
\end{figure}
Not forgetting any good policy learned before is a key concept of the eco-system approach, which is confirmed by the catastrophic forgetting avoidance  index $\xi$ (Figure \ref{easy_generalization_forget}). The single agent approach forgets environments previously learned while learning new environments and the eco-system is always able to solve any environment previously solved.
The PPO algorithm with stable-baselines3 framework gives similar results, even if the performance of the single agent is lower than when using DDQN. (Figure \ref{easy_generalization_gen} and \ref{easy_generalization_forget}).
\paragraph{Submarine-hard}The second experiment was conducted on the harder to generalize submarine environment (Submarine-hard).
The eco-system achieves similar performance on the generalizability index based on threshold $\omega$ (Figure \ref{hard_generalization_gen}) on new environments while the performance of the single agent decreases.
\begin{figure}[ht]
	\centering
	\includegraphics[width=5cm]{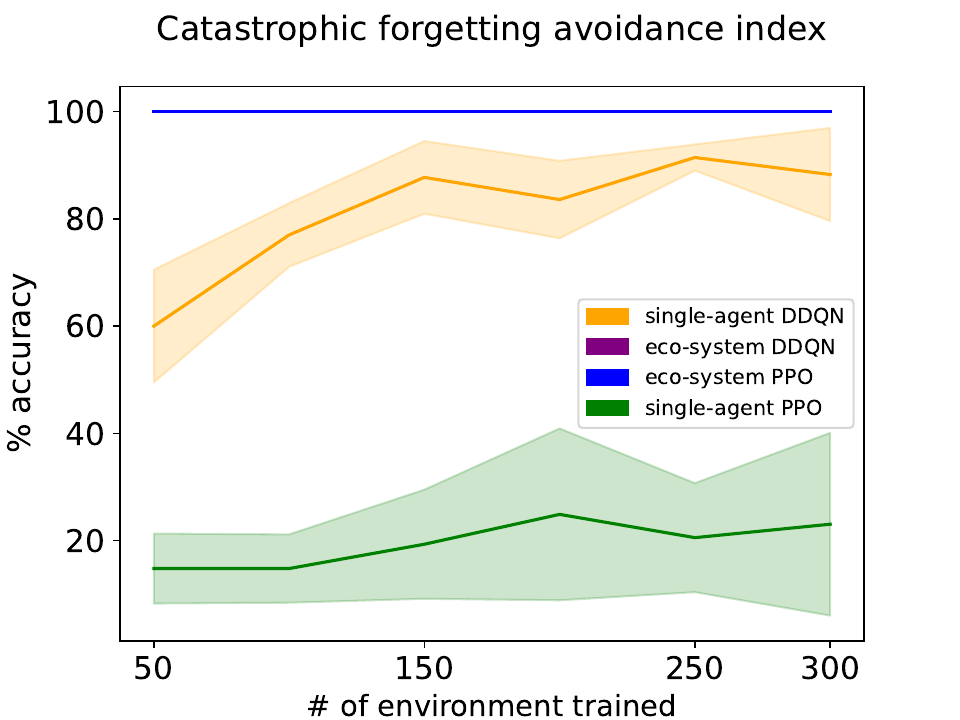}
	\caption{Catastrophic forgetting Submarine hard}
	\label{hard_generalization_forget}
\end{figure}
The catastrophic forgetting avoidance index $\xi$ (Figure \ref{hard_generalization_forget}) explains the results seen with the other chart, the catastrophic forgetting effect creates a lot of instability for the single agent. The agent is forgetting a significant part of what it has learned before. 
This is typical, as the agent aims to solve the current environment, it adapts its neural network to solve specific states, which means over-fitting to this environment. The likelihood to find similar states on another environment leading to the same action is really low, which means that the single agent has to modify its neural network to adapt to the new states and actions.
We also run the same experiments using the PPO algorithm with stable-baselines3 framework, and got results confirming the superiority of the eco-system approach (Figure \ref{hard_generalization_gen} and \ref{hard_generalization_forget}).
On both experiments we can see that the better performance in generalizability and the removal of catastrophic forgetting comes at the expense of the number of accesses made to the environments by the eco-system, see Figure \ref{hard_access}.
We can however see that when the environment is more challenging for an agent to generalize on, the difference between the single agent and the eco-system in term of accesses is becoming less and less (see Figure \ref{hard_access}).
If we look at the duration to train the single agent and the eco-system on the initial environments for the hard to generalize version (Figure \ref{hard_duration}), it is quite similar. This indicates that even if the eco-system requires more access to the environment, most of them are made as inference and not as learning, which has a lower impact on the overall duration, where the single-agent focus most of the access on learning and Neural Network back-propagation.
\subsection{Minigrid}
\begin{figure}[ht]
	\centering
	\includegraphics[width=5cm]{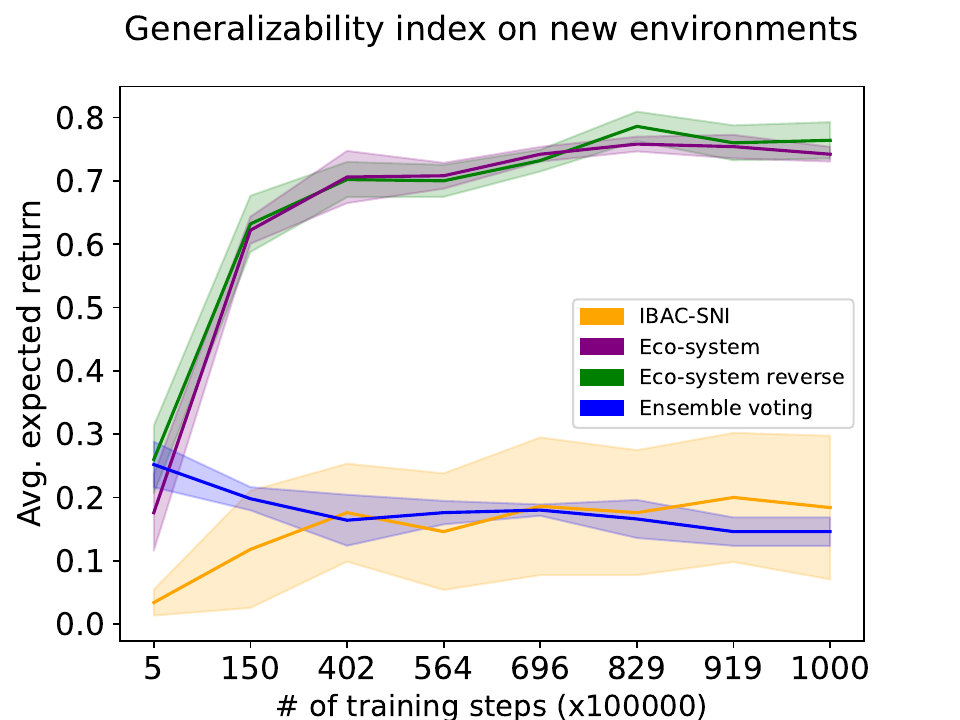}
	\caption{Benchmark on Minigrid FourRoom}
	\label{benchmark_minigrid_fourroom}
\end{figure}

\begin{figure}[ht]
	\centering
	\includegraphics[width=5cm]{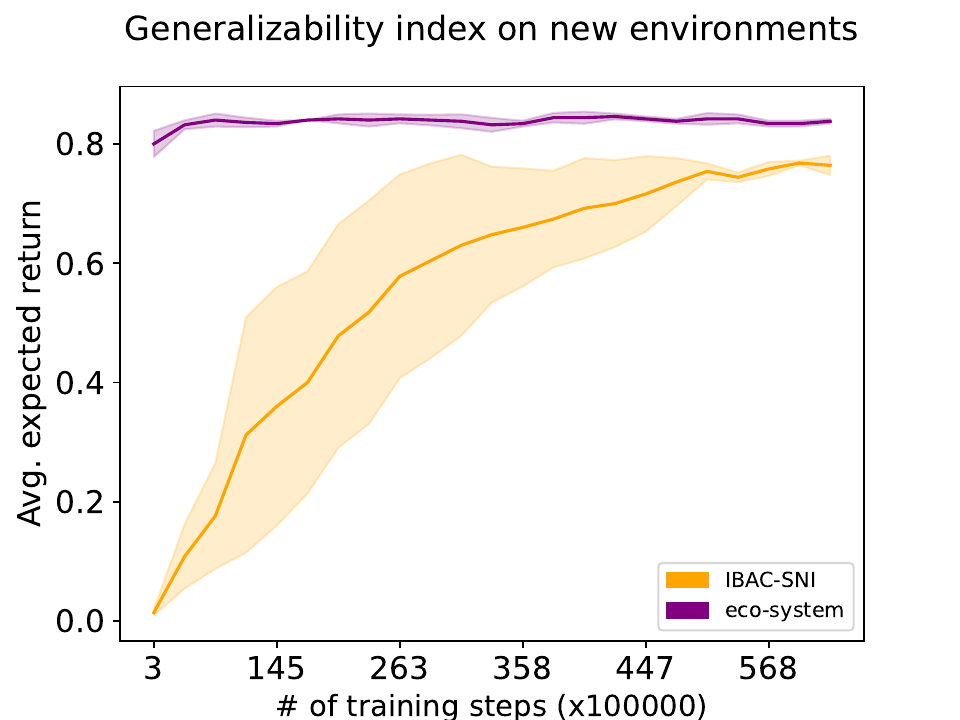}
	\caption{Benchmark on Minigrid Multiroom}
	\label{benchmark_minigrid_multiroom}
\end{figure}

\paragraph{Minigrid FourRoom} The next experiment was run on the Minigrid FourRoom environment to compare how the eco-system performs in comparison with the proposed approach by Igl et al. \cite{Migl2019}.
 The generalization assessment for the FourRoom and MultiRoom environments is based on the average of total return $\zeta$ to match the metrics used in the benchmark approach (cf. \cite[Igl et al., 2019] {Migl2019}), which is independent of the threshold. The threshold is only used to end the training of the new agent on a given environment before it gets added to the pool.
The FourRoom environment is hard to generalize on, as a lot of parameters are randomly chosen: the start position of the agent, the position of the goal to reach, and the opening in the walls).
The eco-system performs better than IBAC-SNI algorithm used in the benchmarking paper (\cite [Igl et al., 2019]{Migl2019}) on new randomly chosen environments (see Figure \ref{benchmark_minigrid_fourroom} and \ref{benchmark_minigrid_multiroom}) after being trained on $10^{8}$ steps on other randomly chosen environments.
This confirms the results gathered on the submarine environment, when the generalization is difficult for a single agent, the eco-systems performs significantly better when compared to a strong baseline, also reaching quicker an higher average total reward.
The standard deviation of the eco-system results is also smaller than the one of the benchmark approach, showing it is far more stable.
The eco-system also outperforms the ensemble voting technique described by Wiering et al. \cite{MAWiering2008} (cf. Figure \ref{benchmark_minigrid_fourroom})
We can also notice that the ensemble voting technique seems to see a decrease in its performance when more agents are added. Having the choice of the action to be taken linked with the answer of the majority of agents can create cases where the agents in the pool vote for a bad action if only a small number of agents were able to properly solve the environment.
\subsection{Impact of pool ordering}
We also study the impact of the ordering function applied to the pool of agents.
In the standard version, the agent that has solved the highest number of environments is tested first when looking at a new environment, based on the fact that an agent scoring high on generalization may have a better chance to solve the new environment. Another approach is to consider that the more environments an agent can solve, the more we may reach the limit of the neural network to solve new environments and that ordering by the reverse order may bring more performance.
The results show that this choice has no impact on the performance of the eco-system approach (see Figure \ref{benchmark_minigrid_fourroom}). 
\subsection{Impact of threshold}
\begin{figure}[ht]
	\centering
	\includegraphics[width=5cm]{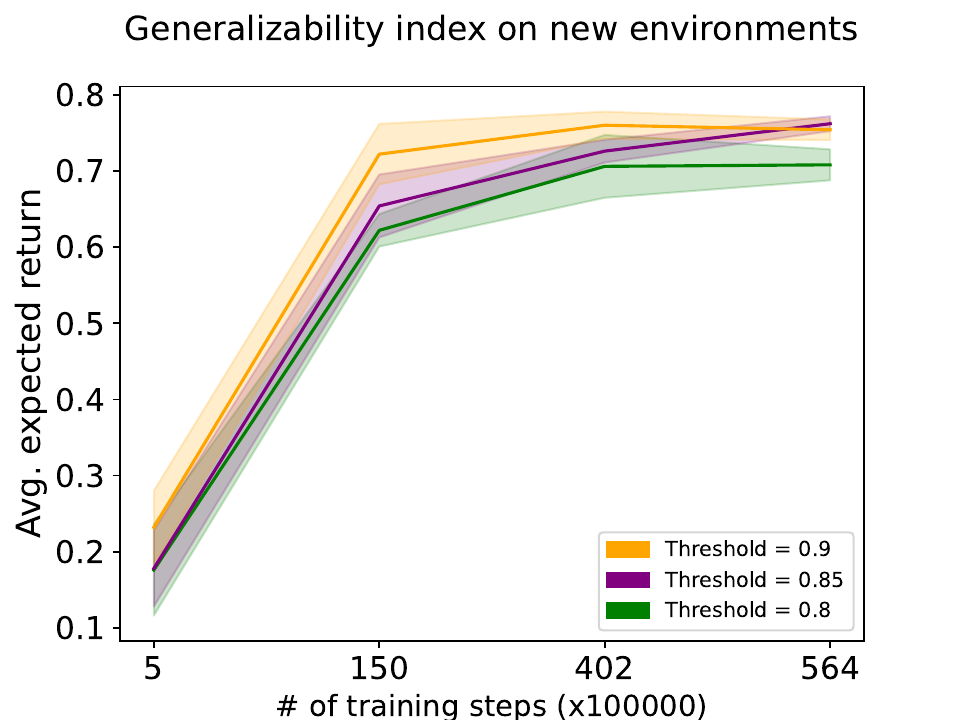}
	\caption{Impact of the threshold on Minigrid FourRoom environment}
	\label{threshold_change}
\end{figure}

In order to find the proper threshold for the Minigrid environment, the following experiment has been run using 3 options : 0.8, 0.85 and 0.9
 (see Figure \ref{threshold_change}).
The results of the experiment indicate that changing the threshold does not have a significant impact on the result of the experiments.
The modification of the threshold has an impact on the number of agents needed in the pool to reach the same result. We have then made the choice to use 0.8 as it was the one using the minimum number of agents and offering a better generalizability by agent.
The following results show the number of agents in the pool after training the eco-system on 150 environments.
\begin{center}
\begin{tabular}{ |c||c|  }
 \hline
 Threshold& Number of agents\\
 \hline
 0.8 & 79\\
 0.85 & 90\\
 0.9 & 107\\
 \hline
\end{tabular}
\end{center}
This shows that our approach is very robust and not impacted by modifications / adjustments of some of its core parameters.
\subsection{Minigrid MultiRoom} The last experiment was run on the Minigrid MultiRoom environment used by Igl et al. \cite{Migl2019}.
This experiment allows to benchmark our approach against a state-of-the-art published approach by using the exact same environment and performance indicator. 
Our approach and the IBAC-SNI approach are quite different, the eco-system focus on solving a given environment before moving to another one while IBAC-SNI trains on multiple environments without trying to fully solve one, which explains why after a similar number of training steps, the eco-system offers better generalization performance. 
As seen on Figure \ref{benchmark_minigrid_multiroom}, the eco-system consistently outperforms the IBAC-SNI approach, even if IBAC-SNI gets closer to the eco-system results by the end of the training, which makes sense based on how learning is done in each approach as explained above. 
\section{Discussion / conclusion}
\label{discussion_conclusion}
We have introduced a novel approach using an eco-system of agents to improve generalizability of Reinforcement Learning. 
Our approach of an eco-system of agents adapts better (i.e. a higher sum of rewards) to never seen environments and keeps performing better on already learned environments, in all configurations explored in this paper, than a single agent or ensemble techniques.
The experiments also show that our approach eliminates catastrophic forgetting when adapting to new environments.
When a given environment is learned, it will never be forgotten. The agent solving it from the pool may change, but there will always be at least one agent from the pool that can solve it.
This is a key point of our solution, which makes it a perfect fit for setup where forgetting a previously trained environment is not an option.
We have also showed that our proposed solution performs at least as well as the other solutions explored in this paper, and that it is less prone to generalizability level drop or training time increase than the other solutions as the generalization to new environments becomes harder.
Our approach shows a greater stability in terms of generalization to new environments when the number of environments on which the system has been trained increases compared to other approaches.
Future work will be focused on trying to optimize the initialization of the agents in the pool and ensuring a better generalization capabilities to the newly created agents. 


\begin{thebibliography}{9}
	\bibitem{KCobbe2019}
    [Cobbe et al., 2019] Cobbe, K., Klimov, O., Hesse, C., Kim, T. \& Schulman, J.. (2019). \textit{Quantifying Generalization in Reinforcement Learning.}  https://proceedings.mlr.press/v97/cobbe19a.html
    \bibitem{JFarebrother2018}
    [Farebrother et al., 2018] Farebrother, J., Machado, M. C., \& Bowling, M. (2018). \textit{Generalization and regularization in DQN.} arXiv preprint arXiv:1810.00123.
    \bibitem{NJustesen2018}
    [Justesen et al., 2018]
    Justesen, N., Torrado, R. R., Bontrager, P., Khalifa, A., Togelius, J., \& Risi, S. (2018). \textit{Illuminating generalization in deep reinforcement learning through procedural level generation.} arXiv preprint arXiv:1806.10729.
    \bibitem{SGhissian2019}
    [Ghiassian et al., 2019]
    Ghiassian, S., Rafiee, B., Lo, Y. L., \& White, A. (2020). \textit{Improving performance in reinforcement learning by breaking generalization in neural networks.} arXiv preprint arXiv:2003.07417.
    \bibitem{HVanHasselt2019}
    [Van Hasselt et al., 2019]
    Van Hasselt, H., Guez, A., \& Silver, D. (2016, March). \textit{Deep reinforcement learning with double q-learning.} In Proceedings of the AAAI conference on artificial intelligence (Vol. 30, No. 1).
	\bibitem{ANichol2018}
	[Nichol et al., 2018]
    Nichol, A., Pfau, V., Hesse, C., Klimov, O., \& Schulman, J. (2018). \textit{Gotta learn fast: A new benchmark for generalization in rl.} arXiv preprint arXiv:1804.03720.
	\bibitem{CPacker2018}
	[Packer et al., 2018]
    Packer, C., Gao, K., Kos, J., Krähenbühl, P., Koltun, V., \& Song, D. (2018). \textit{Assessing generalization in deep reinforcement learning.} arXiv preprint arXiv:1810.12282.
    \bibitem{VFrancoisLavet2016}
	[Fran\c{c}ois-Lavet et al., 2016]
	V. Fran\c{c}ois-Lavet et al. (2016).\textit{DeeR.} Available from https://deer.readthedocs.io/.
	\bibitem{stable-baselines3}
	[Raffin et al., 2019]
	A.Raffin, A.Hill, M.Ernestus, A.Gleave, A.Kanervisto, Anssi, \& N.Dormann. (2019). \textit{Stable Baselines3.} Available from https://github.com/DLR-RM/stable-baselines3
	\bibitem{VFrancoisLavet2018}
    [Fran\c{c}ois-Lavet et al., 2018]
    François-Lavet, V., Henderson, P., Islam, R., Bellemare, M. G., \& Pineau, J. (2018). \textit{An introduction to deep reinforcement learning.} arXiv preprint arXiv:1811.12560.
	\bibitem{gym_minigrid}
	[Chevalier-Boisvert et al., 2018]
	M. Chevalier-Boisvert, L. Willems, \& S. Pal. (2018). \textit{Minimalistic Gridworld Environment for OpenAI Gym.} Available from https://github.com/maximecb/gym-minigrid
	\bibitem{Migl2019}
	[Igl et al., 2019]
    Igl, M., Ciosek, K., Li, Y., Tschiatschek, S., Zhang, C., Devlin, S., \& Hofmann, K. (2019). \textit{Generalization in reinforcement learning with selective noise injection and information bottleneck.} arXiv preprint arXiv:1910.12911.
	\bibitem{asonar2020}
	[Sonar et al., 2021]
    Sonar, A., Pacelli, V., \& Majumdar, A. (2021, May). \textit{Invariant policy optimization: Towards stronger generalization in reinforcement learning.} In Learning for Dynamics and Control (pp. 21-33). PMLR.
	\bibitem{salver2020}
	[Alver et al., 2020]
    Alver, S., \& Precup, D. (2020). \textit{A brief look at generalization in visual meta-reinforcement learning.} arXiv preprint arXiv:2006.07262.
    \bibitem{yduan2016}
    [Duan et al.,2016]
    Duan, Y., Schulman, J., Chen, X., Bartlett, P. L., Sutskever, I., \& Abbeel, P. (2016). \textit{Rl2: Fast reinforcement learning via slow reinforcement learning.} arXiv preprint arXiv:1611.02779.
	\bibitem{jzchen2020}
	[Chen, 2020]
    Chen, J. Z. (2020). \textit{Reinforcement Learning Generalization with Surprise Minimization.} arXiv preprint arXiv:2004.12399.
	\bibitem{Xlu2020}
	[Lu et al., 2020]
    Lu, X., Lee, K., Abbeel, P., \& Tiomkin, S. (2020). \textit{Dynamics Generalization via Information Bottleneck in Deep Reinforcement Learning.} arXiv preprint arXiv:2008.00614
	\bibitem{rssutton1998}
	[Sutton and Barto, 1998]
    Sutton, R. S., \& Barto, A. G. (1998). \textit{Introduction to reinforcement learning} (Vol. 135) Cambridge: MIT press.
	\bibitem{jschulman2017}
	[Schulman et al., 2017]
    Schulman, J., Wolski, F., Dhariwal, P., Radford, A., \& Klimov, O. (2017). \textit{Proximal policy optimization algorithms.} arXiv preprint arXiv:1707.06347.
	\bibitem{mawiering2012}
	[Wiering et Van Otterlo, 2012]
    Wiering, M. A., \& Van Otterlo, M. (2012). \textit{Reinforcement learning. Adaptation, learning, and optimization.}
	\bibitem{catkinson2020}
	[Atkinson et al., 2021]
    Atkinson, C., McCane, B., Szymanski, L., \& Robins, A. (2021). \textit{Pseudo-rehearsal: Achieving deep reinforcement learning without catastrophic forgetting.} Neurocomputing 428, 291-307.
	\bibitem{JKickpatrick2016}
	[Kirkpatrick et al., 2016]
    Kirkpatrick, J., Pascanu, R., Rabinowitz, N., Veness, J., Desjardins, G., Rusu, A. A., ... \& Hadsell, R. (2017). \textit{Overcoming catastrophic forgetting in neural networks.} Proceedings of the national academy of sciences 114(13), 3521-3526.
	\bibitem{AARusu2016}
	[Rusu et al., 2016]
    Rusu, A. A., Rabinowitz, N. C., Desjardins, G., Soyer, H., Kirkpatrick, J., Kavukcuoglu, K., ... \& Hadsell, R. (2016). \textit{Progressive neural networks.} arXiv preprint arXiv:1606.04671.
	\bibitem{sb3PPO}
	[OpenAI]
	\textit{Stable baselines 3. Details on stable baselines 3 PPO implementation.} Available from
	https://spinningup.openai.com/en/latest/algorithms/ppo.html.
    \bibitem{MAWiering2008}
    [Wiering et al., 2008]
    Marco A Wiering and H. V. Hasselt
    \textit{Ensemble Algorithms in Reinforcement Learning} IEEE Transactions on Systems, Man, and Cybernetics, Part B (Cybernetics)
\end{thebibliography}
\end{document}

\maketitle
\section{Experiments results}
\subsection{Indicators used}
For the submarine-easy and submarine-hard experiments, the performance has been evaluated using the adaptability index based on a threshold and the catastrophic forgetting avoidance index. The number of access to the environment as well as the training duration have also been gathered as additional metrics.
For the minigrid experiments, the performance was evaluated on the adaptability index based on the average expected return obtained.
This change of calculation for the minigrid experiment has been made to match the benchmarking IBAC-SNI paper in order to avoid any modification on their code.
\subsection{Charts presented}
We have run each experiments 5 times and we are showing in the charts the mean results over these five runs (line) as well as the corresponding standard deviation.
In order to match the code defined in the benchmarking IBAC-SNI paper, the metrics used for the experiments on minigrid environment are indicated as steps and not the number of environments on which both approaches have been trained.
\section{Submarine experiments implementation}
The submarine experiments implementation (submarine-easy and submarine-hard) use the DeeR and Stable Baselines 3 frameworks to learn and solve the environments.
The DeeR framework is used with the DDQN algorithm, when Stable Baselines 3 is used with PPO.
\subsection{Network architecture}
The state given by the submarine environment is a list of values, composed of the flatten array [x:x+5,0:11] with the addition of the value of x at the end for submarine-easy environment. For submarine hard environment it is composed of the flatten array [x:x+15,0:11] with the addition of the value of x and y at the end.
The network used by DDQN is structured as follow:
\begin{itemize}
    \item one dense layer of size 50 with Relu activation
    \item one dense layer of size 20 with Relu activation
    \item the output layer of size 3 (one per action)
\end{itemize}
The Network architecture used by PPO is based on the MlpPolicy from Stable Baselines3, and is defined in the back-end by the framework itself, the configuration at our level was limited to the Hyper-parameters which were let to their default values.
\subsection{Hyper-parameters}
In order to train the agents using the DeeR framework with DDQN, the following hyper-parameters have been used:
\begin{itemize}
    \item Learning rate = 0.005
    \item Discount factor = 0.9 increasing to 0.99
    \item Epsilon = 1 decreasing to 0.1
    \item Replay memory = 1000000
\end{itemize}
In order to train the agents using the Stable baselines3 framework with PPO, the following hyper-parameters have been used:
\begin{itemize}
    \item Policy = MlpPolicy
    \item Learning rate = 0.0003
    \item Discount factor = 0.99
    \item Clip range = 0.2
\end{itemize}
\section{Minigrid experiments implementation}
The minigrid experiments implementation (FourRooms and MultiRooms) use the Stable Baselines3 framework to learn and solve the environments. This framework is used with the PPO algorithm.
The state used is the standard RGB output generated by Minigrid.
\subsection{Network architecture}
The Network architecture is based on the MlpPolicy from Stable Baselines3, and is defined in the back-end by the framework itself, the configuration at our level was limited to the Hyper-parameters which were let to their default values.
\subsection{Hyper-parameters}
In order to train the agents using the Stable baselines3 framework with PPO, the following hyper-parameters have been used:
\begin{itemize}
    \item Policy = MlpPolicy
    \item Learning rate = 0.0003
    \item Discount factor = 0.99
    \item Clip range = 0.2
\end{itemize}
\section{Code}
\subsection{Agent code}
The code detailed here is limited to the two functions in charge of training an agent on a given environment and checking if an agent can solve an environment.
The first function is in charge to check the expected return which can be gathered by the agent when running on a given environment.
\begin{algorithm}[H]
	\caption{$Agent(DeeR) - check(env)$} 
    \begin{algorithmic}
    \State $env.reset()$
    \State $totalReward = 0$
    \While{$env.inTerminalState() == False$}
        \State $action = agent.testPolicy.action(env.observe())$
        \State $reward = env.act(action)$
        \State $totalReward = totalReward + reward$
    \EndWhile
    \State $return(totalReward)$
    \end{algorithmic}
\end{algorithm}
\begin{algorithm}[H]
	\caption{$Agent(Stable Baselines3) - check(env)$} 
    \begin{algorithmic}
    \State $state = env.reset()$
    \State $totalReward = 0$
    \State $done = False$
    \While{$done == False$}
        \State $action = model.predict(state)$
        \State $state,reward,done = env.step(action)$
        \State $totalReward = totalReward + reward$
    \EndWhile
    \State $return(totalReward)$
    \end{algorithmic}
\end{algorithm}
The second function is in charge to train the agent until it reaches a given threshold on an environment.
\begin{algorithm}[H]
	\caption{$Agent(DeeR) - solve(env)$} 
    \begin{algorithmic}
    \State $solved = False$
    \While{$solved == False$}
        \State $env.reset()$
        \State $agent.runEpoch(1)$
        \If {$check(env)>=100$}
            \State $solved = True$
        \EndIf
    \EndWhile
    \end{algorithmic}
\end{algorithm}

\begin{algorithm}[H]
	\caption{$Agent(Stable Baselines3) - solve(env)$} 
    \begin{algorithmic}
    \State $solved = False$
    \While{$solved == False$}
        \State $model.learn(totalTimesteps=100000)$
        \If {$check(env)>=0.8$}
            \State $solved = True$
        \EndIf
    \EndWhile
    \end{algorithmic}
\end{algorithm}

\subsection{Eco-system code}
The code detailed here is limited to the two methods in charge of training the eco-system on a new environment or checking if the eco-system can solve a given environment.
These functions are the same for both the DeeR and Stable baselines3 implementation as they are reusing the functions defined at the agent level and not interacting with the framework itself.
The only change between the two environments is the modification of the success threshold, 100 for submarine environments and 0.8 for minigrid.
The first function is in charge to check the expected return which can be gathered by the eco-system when running on a given environment. This function use the check function from each agents in the eco-system to determine the best score available.
\begin{algorithm}[H]
	\caption{$Eco-system - test(env)$} 
    \begin{algorithmic}
    \State $maxRes=0$
    \State $found = False$
    \State $j=0$
    \While{$found == False\ and\ j<len(AgentArray)$}
        \If{$env in agentArray[j].solvedEnv$}
            \State $found = True$
            \State $maxRes=agentArray[j].check(env)$
        \EndIf
        \State $j=j+1$
    \EndWhile
    \State $j=0$
    \While{$found == False\ and\ j<len(AgentArray)$}
        \State $res = agentArray[j].check(env)$
        \If{$res > maxRes$}
            \State $max_res = res$
        \EndIf
        \If{$res >= 0.8$}
            \State $found = True$
        \EndIf
        \State $j=j+1$
    \EndWhile
    \State $return(maxRes)$
\end{algorithmic}
\end{algorithm}
The second function is in charge to check if the eco-system can solve an environment with its existing agents in its pool or to add a new agent and train in on the environment. 
This function use the check function of all agents as well as the agent solve function when a new agent should be added to the pool and trained.
This function has been defined in the main paper in a more mathematical way, the version below is closer to the code written to run the experiments.

\begin{algorithm}[H]
	\caption{$Eco-system - train(env)$} 
    \begin{algorithmic}
    \State $solved = False$
	\State $i = 0$
	\While{$solved == False\ and\ i<len(agentArray)$}
		\State $agTest =agentArray[i]$
    	\If { $agTest.check(env)>=0.8$}
			\State $solved = True$
		\Else
			\State $i = i + 1$
		\EndIf
	\EndWhile
	\If{$solved == False$}
		\State $ag= newAgent()$
		\State $ag.solve()$
		\State $agentArray.add(ag)$
		\State $i = len(agentArray)-1$
		\State $agentArray[i].addEnv(env)$
        \State $k = 0$
		\While {$k<len(agentArray)$}
			\State $listEnv =agentArray[k].listEnv$
	        \State $canSolveAll=True$
			\For {$envTest\ in\ listEnv$}					  	
			    \If {$agentArray[i].check(envTest)>=0.80$}
					\If {$envTest\ not\ in\ agentArray[i].solvedEnv$}
						\State $agentArray[i].addEnv(envTest)$
					\EndIf
				\Else
					\State $canSolveAll=False$
				\EndIf
			\EndFor
			\If{$canSolveAll==True\ and\ k!=i$}
				\State $agentArray.removeAgent(k)$
				\If{$k<i$}
					\State $i= i - 1$
				\EndIf
			\Else
				\State $k = k +1$
			\EndIf
		\EndWhile
	\Else
		\If {$env\ not\ in\ agentArray[i].solvedEnv$}
		    \State $agentArray[i].addEnv(env)$
        \EndIf
    \EndIf
    \State $agentArray.sortbyEnv(Descending)$
\end{algorithmic}
\end{algorithm}